# Classification of Images Using Support Vector Machines

*Gidudu Anthony, * Hulley Greg and *Marwala Tshilidzi
*Department of Electrical and Information Engineering, University of the Witwatersrand,
Johannesburg, Private Bag X3, Wits, 2050, South Africa
Anthony.Gidudu@wits.ac.za, greghul@icon.co.za, Tshilidzi.Marwala@wits.ac.za

**Abstract:** Support Vector Machines (SVMs) are a relatively new supervised classification technique to the land cover mapping community. They have their roots in Statistical Learning Theory and have gained prominence because they are robust, accurate and are effective even when using a small training sample. By their nature SVMs are essentially binary classifiers, however, they can be adopted to handle the multiple classification tasks common in remote sensing studies. The two approaches commonly used are the One-Against-One (1A1) and One-Against-All (1AA) techniques. In this paper, these approaches are evaluated in as far as their impact and implication for land cover mapping. The main finding from this research is that whereas the 1AA technique is more predisposed to yielding unclassified and mixed pixels, the resulting classification accuracy is not significantly different from 1A1 approach. It is the authors conclusions that ultimately the choice of technique adopted boils down to personal preference and the uniqueness of the dataset at hand.

## 1.0   INTRODUCTION

Over the last three decades or so, remote sensing has increasingly become a prime source of land cover information (Kramer, 2002; Foody and Mathur, 2004a). This has been made possible by advancements in satellite sensor technology thus enabling the acquisition of land cover information over large areas at various spatial, temporal spectral and radiometric resolutions. The process of relating pixels in a satellite image to known land cover is called image classification and the algorithms used to effect the classification process are called image classifiers (Mather, 1987). The extraction of land cover information from satellite images using image classifiers has been the subject of intense interest and research in the remote sensing community (Foody and Mather, 2004b). Some of the traditional classifiers that have been in use in remote sensing studies include the maximum likelihood, minimum distance to means and the box classifier. As technology has advanced, new classification algorithms have become part of the main stream image classifiers such as decision trees and artificial neural networks. Studies have been made to compare these new techniques with the traditional ones and they have been observed to post improved classification accuracies (Peddle et al. 1994; Rogan et al. 2002; Li et al. 2003; Mahesh and Mather, 2003). In spite of this, there is still considerable scope for research for further increases in accuracy to be obtained and a strong desire to maximize the degree of land cover information extraction from remotely sensed data (Foody and Mather, 2004b). Thus, research into new methods of classification has continued, and support vector machines (SVMs) have recently attracted the attention of the remote sensing community (Huang et al., 2002).

Support Vector Machines (SVMs) have their roots in Statistical Learning Theory (Vapnik, 1995). They have been widely applied to machine vision fields such as character, handwriting digit and text recognition (Vapnik, 1995; Joachims, 1998), and more recently to satellite image classification (Huang et al, 2002; Mahesh and Mather, 2003). SVMs, like Artificial Neural Networks and other nonparametric classifiers have a reputation for being robust (Foody and Mathur, 2004a; Foody and Mathur, 2004b). SVMs function by nonlinearly projecting the training data in the input space to a feature space of higher (infinite) dimension by use of a kernel function. This results in a linearly separable dataset that can be separated by a linear classifier. This process enables the classification of remote sensing datasets which are usually nonlinearly separable in the input space. In many instances, classification in high dimension feature spaces results in over-fitting in the input space, however, in SVMs over-fitting is controlled through the principle of structural risk minimization (Vapnik, 1995). The empirical risk of misclassification is minimised by maximizing the margin between the data points and the decision boundary (Mashao, 2004). In practice this criterion is softened to the minimisation of a cost factor involving both the complexity of the classifier and the degree to which marginal points are misclassified. The tradeoff between these factors is managed through a margin of error parameter (usually designated C) which is tuned through cross-validation procedures (Mashao, 2004). The functions used to project the data from input space to feature space are sometimes called kernels (or kernel machines), examples of which include polynomial, Gaussian (more commonly referred to as radial basis functions) and quadratic functions. Each function has unique parameters which have to be determined prior to classification and they are also usually determined through a cross validation process. A deeper mathematical treatise of SVMs can be found in Christianini (2002), Campbell (2000) and Vapnik (1995).

By their nature SVMs are intrinsically binary classifiers (Melgani and Bruzzone, 2004) however there exist strategies by which they can be adopted to multiclass tasks associated with remote sensing studies. Two of the common approaches are the One-Against-One (1A1) and One-Against-All (1AA) techniques. This paper seeks to explore these two approaches with a view of discussing their implications for the classification of remotely sensed images.

## 2.0 SVM MULTICLASS STRATEGIES

As mentioned before, SVM classification is essentially a binary (two-class) classification technique, which has to be modified to handle the multiclass tasks in real world situations e.g. derivation of land cover information from satellite images. Two of the common methods to enable this adaptation include the 1A1 and 1AA techniques. The 1AA approach represents the earliest and most common SVM multiclass approach (Melgani and Bruzzone, 2004) and involves the division of an N class dataset into N two-class cases. If say the classes of interest in a satellite image include water, vegetation and built up areas, classification would be effected by classifying water against non-water areas i.e. (vegetation and built up areas) or vegetation against non-vegetative areas i.e. (water and built up areas). The 1A1 approach on the other hand involves constructing a machine for each pair of classes resulting in N(N-1)/2 machines. When applied to a test point, each classification gives one vote to the winning class and the point is labeled with the class having most votes. This approach can be further modified to give weighting to the voting process. From machine learning theory, it is acknowledged that the disadvantage the 1AA approach has over 1A1 is that its performance can be compromised due to unbalanced training datasets (Gualtieri and Cromp, 1998), however, the 1A1 approach is more computationally intensive since the results of more SVM pairs ought to be computed. In this paper, the performance of these two techniques are compared and evaluated to establish their performance on the extraction of land cover information from satellite images.

## 3.0 METHODOLOGY

The study area was extracted from a 2001 Landsat scene (row 171 and row 60). It is located at the source of River Nile in Jinja, Uganda. The bands used in this research consisted of Landsat's optical bands i.e. bands 1, 2, 3, 4, 5 and 7. The classes of interest were built up area, vegetation and water. IDRISI Andes was used for preliminary data preparation such as sectioning out of the study area from the whole scene and identification of training data. This data was then exported into a form readable by MATLAB (Version 7) for further processing and to effect the classification process. The SVMs that were used included the Linear, Polynomial, Quadratic and Radio Basis Function (RBF) SVMs. Each classifier was employed to carry out 1AA and 1A1 classification. The classification results for both 1AA and 1A1 were then imported into IDRISI for georeferencing, GIS integration, accuracy assessment and derivation of land cover maps. The following four parameters formed the basis upon which the two multiclassification approaches were compared: Number of unclassified pixels, number of mixed pixels, final accuracy assessment and the 95% level of significance of the difference between overall accuracies of the two approaches (i.e. $|Z| > 1.96$).

## 4.0 RESULTS, DISCUSSION AND CONCLUSION

Table 1 gives a summary of the unclassified and mixed pixels resulting from 1A1 and 1AA classification. From Table 1 it is evident that the 1AA approach to multiclass classification has exhibited a higher propensity for unclassified and mixed pixels than the 1A1 approach. A graphical consequence of this is evident in the derived land cover maps shown in Figures 1 – 4. Figures 1a, 2a, 3a and 4a are a result of adopting 1A1, while Figures 1b, 2b, 3b and 4b depict classification output following the use of the 1AA approach. The 'sputtering' of black represent the unclassified pixels while that of red shows pixels belonging to more than one class. The nature of the 1AA is such that the exact class of the mixed pixels cannot be determined and for accurate analysis all such pixels ought to be grouped together. From Table 1 and the corresponding derived land cover maps, it is clear that the 1A1 has posted more aesthetic results.

**Table 1:** Summary of number of unclassified and mixed pixels

| Classifier | Type | 1A1 | 1AA |
|---|---|---|---|
| Linear | Unclassified Pixels | 16 | 700 |
| | Mixed Pixels | 0 | 9048 |
| Quadratic | Unclassified Pixels | 142 | 5952 |
| | Mixed Pixels | 0 | 537 |
| Polynomial | Unclassified Pixels | 69 | 336 |
| | Mixed Pixels | 0 | 2172 |
| RBF | Unclassified Pixels | 103 | 4645 |
| | Mixed Pixels | 0 | 0 |

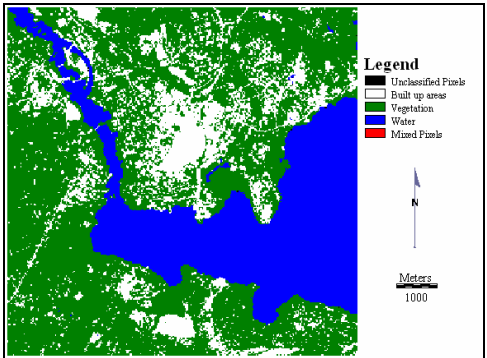
**Figure 1a:** 1A1 - Linear

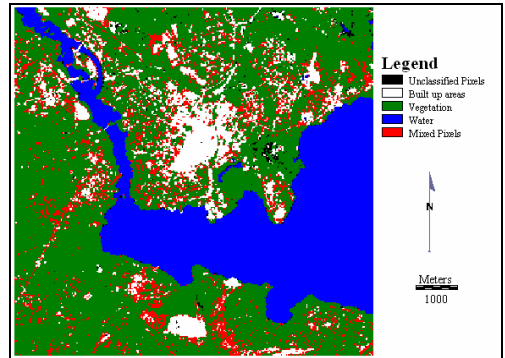
**Figure 1b:** 1AA - Linear

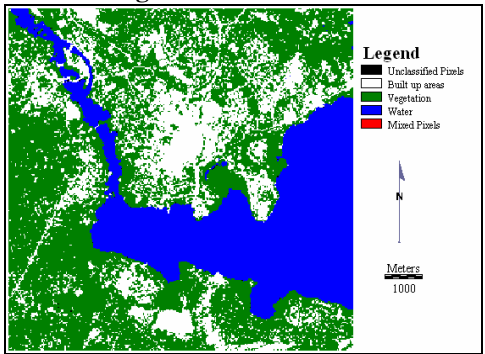
**Figure 2a:** 1A1 - Polynomial

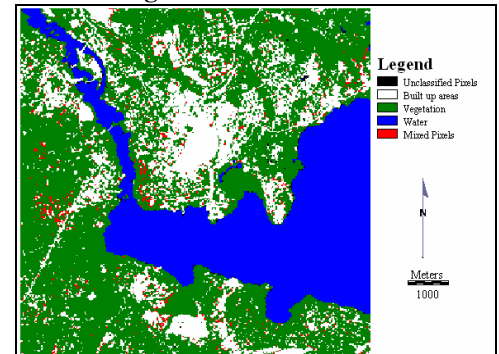
**Figure 2b:** 1AA - Polynomial

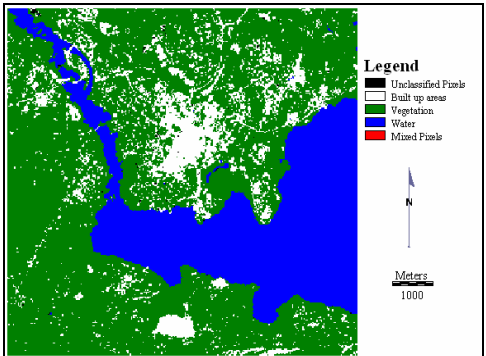
**Figure 3a:** 1A1 - Quadratic

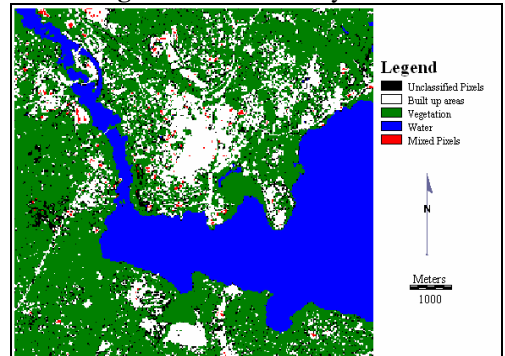
**Figure 3b:** 1AA - Quadratic

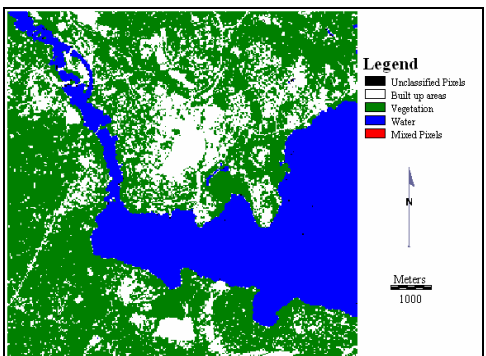
**Figure 4a:** 1A1 - RBF

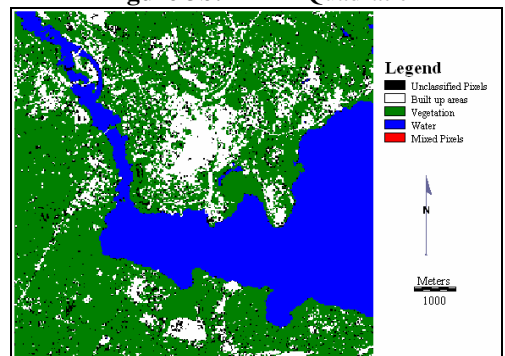
**Figure 4b:** 1AA - RBF

Of greater importance was the effect of the unclassified and mixed pixels on the overall accuracies. Table 2 gives a summary of the kappa accuracies for the various SVM classifiers adopting either of the two approaches.

**Table 2:** Summary of Kappa Values for Corresponding SVMs

| SVM | 1A1 | 1AA | |Z| | Significance |
|---|---|---|---|---|
| Linear | 1.00 | 0.95 | 0.06 | Difference insignificant |
| Quadratic | 0.88 | 0.94 | -0.02 | Difference insignificant |
| Polynomial | 1.00 | 1.00 | 0.00 | No difference |
| RBF | 0.97 | 0.92 | 0.01 | Difference insignificant |

From Table 2, all accuracies would be classified as yielding very strong correlation with ground truth data. The individual performance of the SVM classifiers however show that classification accuracy reduced for the linear and RBF classifiers, stayed the same for the polynomial and increased for the quadratic classifier. Further analysis of these results show that these differences are pretty much insignificant at the 95% confidence interval. It can therefore be concluded that whereas one can be certain of high classification results with the 1A1 approach, the 1AA yields approximately as good classification accuracies. The choice therefore of which approach to adopt henceforth becomes a matter of preference.

## ACKNOWLEDGEMENTS

The authors would like to acknowledge the financial support of the University of the Witwatersrand and National Research Foundation of South Africa